\documentclass[runningheads]{llncs}
\usepackage{siunitx}
\usepackage{graphicx}
\usepackage[table]{xcolor}
\usepackage{comment}
\usepackage{rotating}
\usepackage{paralist}
\usepackage{amsmath}
\usepackage{amssymb}
\usepackage{hyperref}
\usepackage{array}
\usepackage{multirow}
\usepackage{booktabs}
\usepackage{pifont}
\usepackage{ulem}
\usepackage{cleveref}
\usepackage{amsmath}
\usepackage{subcaption}
\usepackage{orcidlink}
\usepackage{makecell}
\usepackage{algorithm}
\usepackage{algpseudocode}

\newcommand{\modelname}{\textsc{LIGHT}}
\newcommand{\tabvspace}{-0.0in}

\begin{document}
\title{LIGHT: Multi-Modal Text Linking on Historical Maps}
%

\author{Yijun Lin\inst{1} \and
Rhett Olson\inst{1} \and
Junhan Wu \inst{1} \and
Yao-Yi Chiang\inst{1} \and
Jerod Weinman \inst{2}
}
\institute{University of Minnesota, Minneapolis, Minnesota, United States \and
           Grinnell College, Grinnell, Iowa, United States
\\ 
\email{\{lin00786, olso9295, wu001412, yaoyi\}@umn.edu, jerod@acm.org}
}
\authorrunning{Lin et al.}

\maketitle

\begin{abstract}
Text on historical maps provides valuable information for studies in history, economics, geography, and other related fields. Unlike structured or semi-structured documents, text on maps varies significantly in orientation, reading order, shape, and placement. Many modern methods can detect and transcribe text regions, but they struggle to effectively ``link'' the recognized text fragments, e.g., determining a multi-word place name.
Existing layout analysis methods model word relationships to improve text understanding in structured documents, but they primarily rely on linguistic features and neglect geometric information, which is essential for handling map text. To address these challenges, we propose \modelname, a novel multi-modal approach that integrates linguistic, image, and geometric features for linking text on historical maps. In particular, \modelname~includes a geometry-aware embedding module that encodes the polygonal coordinates of text regions to capture polygon shapes and their relative spatial positions on an image. \modelname~ unifies this geometric information with the visual and linguistic token embeddings from LayoutLMv3, a pretrained layout analysis model. \modelname~uses the cross-modal information to predict the reading-order successor of each text instance directly with a bi-directional learning strategy that enhances sequence robustness. Experimental results show that \modelname~outperforms existing methods on the ICDAR 2024/2025 MapText Competition data, demonstrating the effectiveness of multi-modal learning for historical map text linking.
\end{abstract}


\setlength{\unitlength}{1in} 
\begin{picture}(0,0) 
    \put(0,7.375){ 
        \parbox[t]{4.25in}{
            {\tiny This version of the contribution has been accepted for publication, after peer review but is not the Version of Record and does not reflect post-acceptance improvements, or any corrections. 
            The Version of Record is available online at: \url{https://doi.org/10.1007/XXXXX}. 
            Use of this Accepted Version is subject to the publisher’s Accepted Manuscript terms of use \url{https://www.springernature.com/gp/open-research/policies/accepted-manuscript-terms}
            }
        }
    }
    \put(0,-2.25){ 
        \parbox[t]{4.25in}{
            {\tiny To appear in \textit{Proceedings of the 19th International
                Conference on Document Analysis and Recognition (ICDAR
                2025)}.  Xu-Cheng Yin, Dimosthenis Karatzas, and Daniel Lopresti (Eds.). Springer Nature LNCS. 2025.
            }
        }
    }
\end{picture}  

\section{Introduction}
Historical maps contain valuable information about geographical, cultural, and environmental changes over time, offering essential insights for research in fields like history, geography, urban planning, cartography, and environmental science~\cite{chiang2020using}. 
Recent advances in map digitization have improved the accessibility of historical maps. For example, automatically extracting text from historical maps unlocks unique insights into historical, political, and cultural contexts, such as place names, facility types, and landmark descriptions~\cite{olson2023automatic}. This process can significantly enhance the ability to search relevant maps through large map archives and facilitate the generation of map metadata~\cite{chiang2023geoai}.
While current text spotting techniques focus on word-level detection and recognition~\cite{zhang2022text,ye2023deepsolo,lin2024hyper}, place names or toponyms often span multiple words.
Therefore, effectively linking these words into location phrases will enhance access to map content and facilitate various downstream tasks, such as georeferencing~\cite{li2020automatic} and entity linking to external databases~\cite{kim2023mapkurator}.



\begin{figure}[ht]
    \centering
    \begin{subfigure}[b]{0.32\textwidth}
        \includegraphics[width=\textwidth]{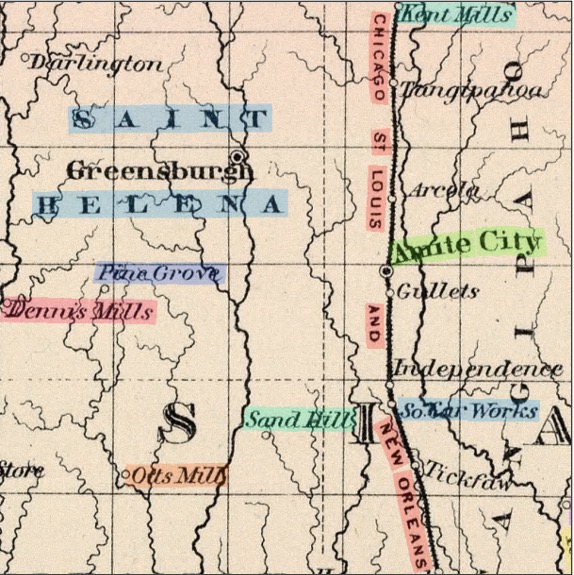}
        \caption{}
        \label{fig:example_c}
    \end{subfigure} 
    \hfill
     \begin{subfigure}[b]{0.32\textwidth}
        \includegraphics[width=\textwidth]{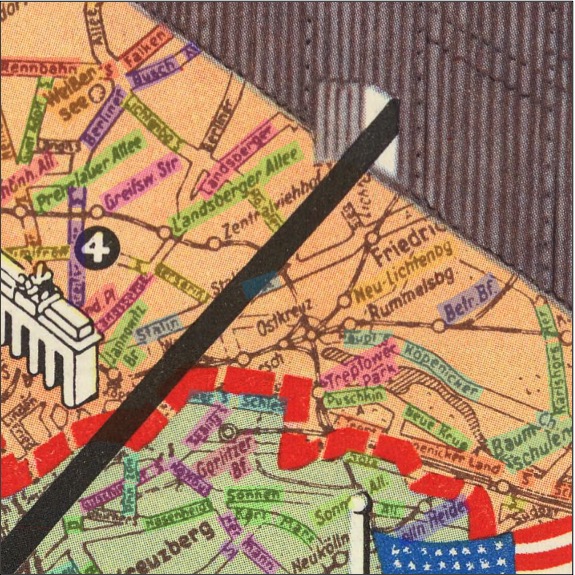}
        \caption{}
        \label{fig:example_a}
    \end{subfigure}
    \hfill
    \begin{subfigure}[b]{0.32\textwidth}
        \includegraphics[width=\textwidth]{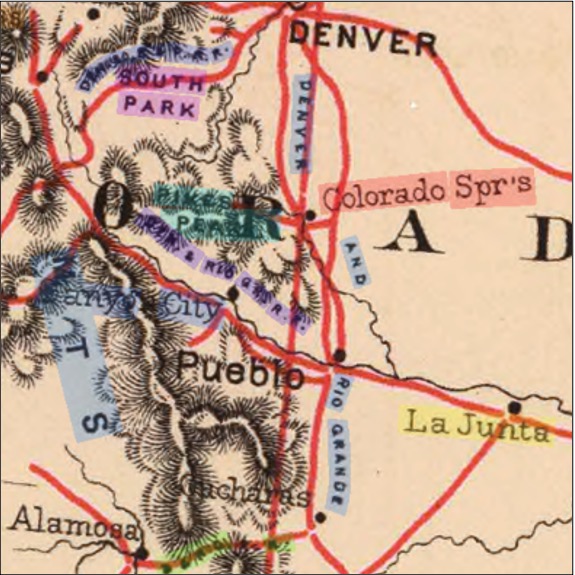}
        \caption{}
        \label{fig:example_b}
    \end{subfigure}
    \caption{Examples of map regions that contain linked words. Words with the same highlighted color belong to the same group ($\geq$2 words).}
    \label{fig:example}
\end{figure}

Linking text (words) in reading order on historical maps is challenging due to their complex and diverse layouts.
Figure~\ref{fig:example} shows three example maps with annotations where a highlighted color indicates a multi-word location phrase.
These annotations present several challenges for automatic linking methods.
For example, a place name might not fall on a single straight line, or unrelated text labels may appear between multi-word phrases, e.g., ``SAINT HELENA'' in Figure~\ref{fig:example_c}.
High-density text regions further complicate distinguishing linked groups, as in Figure~\ref{fig:example_a}.
Additionally, inconsistent layouts across maps can have words far apart representing large regions, e.g., ``DENVER AND RIO GRANDE'' in Figure~\ref{fig:example_b}.
These challenges require text linking methods to effectively capture word relationships from multiple perspectives, including linguistic, visual, and geometric contexts.


Existing map text linking techniques leverage features such as the sizes, heights, and rotation angles of text instances to group related words on maps (e.g.,~\cite{olson2024automatic}).
However, these approaches do not use important text information. Li et al.~\cite{li2020automatic} introduce word embeddings from GloVe~\cite{pennington2014glove} as an additional feature, but the method fails to explicitly capture the linguistic context of words on maps.
In addition, recent advances in document layout analysis, such as DLAFormer~\cite{wang2024dlaformer} and LayoutLMv3~\cite{huang2022layoutlmv3}, can jointly integrate textual and visual features, utilizing multi-modal embeddings and attention mechanisms to understand the structure and semantics of documents.
These models can be leveraged for tasks such as map text linking; however, they are typically pretrained on well-structured documents with uniform text shapes and placements, which might not work well for highly irregular and diverse text layouts on historical maps.
Addressing these issues requires further domain-specific fine-tuning with map data and the integration of geometric information that captures text shapes and spatial relationships between text instances.

\begin{figure}
    \centering
    \includegraphics[width=\textwidth]{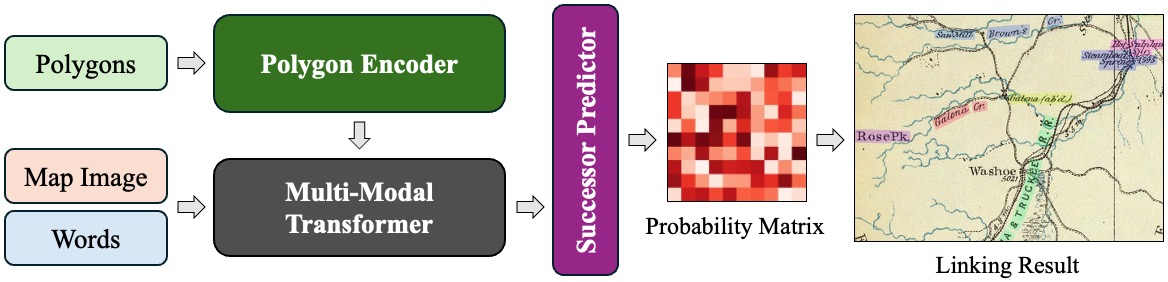}  
    \caption{\modelname~preview: Language, Image, and Geometry for Linking Hierarchical Text on historical maps.}
    \label{fig:preview}
\end{figure}

To address these challenges, we propose a novel multi-modal approach called \modelname, which integrates \underline{\textbf{L}}inguistic, \underline{\textbf{I}}mage/visual, and \underline{\textbf{G}}eometric contexts for linking \underline{\textbf{H}}ierarchical  \underline{\textbf{T}}ext on maps.
As shown in Figure~\ref{fig:preview}, \modelname~first employs a polygon encoder to transform the polygonal coordinates of text regions into fixed-size vector embeddings, capturing irregular polygon shapes and relative positions of text instances.
Then \modelname~utilizes a multi-modal transformer that integrates the transcribed text contents, map image, and polygon embeddings to jointly learn a cross-modal representation for each word. 
Finally, \modelname's successor predictor directly calculates a one-versus-all probability of each word being the successor for a given word, followed by a greedy selection strategy to determine the optimal link sequence.
Additionally, we introduce a bi-directional learning strategy that predicts both the successor and predecessor of each word, enhancing the model's robustness in learning text sequences. 
The main contributions of the paper include,

\begin{itemize}
    \renewcommand\labelitemi{$\bullet$}
    \item We propose a novel multi-modal approach that integrates text content, map image, and text polygon information for text linking on historical maps.
    \item We develop a geometry-aware embedding module that effectively captures polygonal shapes and spatial relationships between text instances, thereby enhancing the model's understanding of text layout on maps.
    \item We introduce a successor prediction mechanism that directly generates text sequences coupled with a bi-directional learning strategy to improve the model's capability of learning coherent text sequences.
\end{itemize}


\section{Related Work}
\subsection{Text Linking on Historical Maps}\label{related: 2.1}
Most existing approaches for text linking on historical maps focus on the downstream task of searching for multi-word place names. The goal is, therefore, to find a given place name containing more than one word from a set of recognized words on a map~\cite{olson2024automatic,kim2023mapkurator,scalora2020changing}. One rule-based method used by the David Rumsey Map Collection~\cite{RumseyCollection} links all pairs of recognized words on maps within a distance of two characters from each other. The width of a character is estimated as the width of a word divided by the number of characters in the word. 
Olson et al.~\cite{olson2024automatic} propose a method to link a given set of recognized words into a minimum spanning tree, by applying Prim's algorithm~\cite{prim1957shortest} to minimize edge costs according to the heuristic. Their heuristic considers factors including spatial proximity, text height, angle of orientation, and capitalization. 
However, these methods tend to draw many more links than necessary and prioritize recall over precision to support multi-word search capabilities. 

There are approaches focusing on directly model the relationships between words on historical maps~\cite{li2020automatic,zou2025recognizing}.
MapKurator~\cite{li2020automatic} combines word embeddings, geometry information, and image features, and uses triplet loss to enforce undirected pairwise linkage. However, mapKurator does not consider word semantics in map context and requires ad-hoc post-processing (e.g., sorting words within a group by coordinates) to generate directional links. 
Self-Sequencer~\cite{zou2025recognizing} adapts a pointer network to sequentially predict successor link connections using only geometry (word boundary points).
In comparison, \modelname~integrates linguistic, visual, and geometric contexts to directly predict the probability of a given word's successor (and predecessor), enhancing robustness and flexibility in text linking on historical maps.

\subsection{Document Layout Analysis}
One line of research in document layout analysis focuses on processing structured (e.g., tables, forms) and semi-structured documents (e.g., receipts, reports)~\cite{powalski2021going,wang2024dlaformer,zhong2023hybrid,huang2022layoutlmv3,appalaraju2021docformer,kim2022ocr}. 
DLAFormer~\cite{wang2024dlaformer} integrates sub-tasks including text region detection, logical role classification, and reading order prediction into a single framework and consolidates these relation prediction labels into a unified label space. 
DINO~\cite{zhong2023hybrid} is a transformer-based framework for document layout analysis but targets large collections of heterogeneous document images. 
Zhang et al.~\cite{zhang2024modeling} propose modeling the document layouts in segment-level reading order using a relation-aware attention module. LayoutLMv3~\cite{huang2022layoutlmv3} employs a multi-modal transformer to integrate text and images.
The pretrained LayoutLMv3 can solve various downstream tasks related to document layout analysis. GeoLayoutLM~\cite{Luo2023GeoLayoutLM} further builds on LayoutLMv3 by explicitly modeling three predefined geometric relations with geometry-related pre-training tasks.

Some graph-based approaches for layout analysis represent document images as graphs, with words or lines as nodes and edges denoting their relationships~\cite{li2018page,li2020page,luo2022doc,wang2022post,wang2024detect}. 
Wang et al.~\cite{wang2024detect} propose a tree-based approach that jointly addresses page object detection, reading order, and hierarchical structure construction. 
SPADE~\cite{hwang2021spatial} tackles spatial dependency parsing by constructing a directed semantic text graph in semi-structured documents. 
These methods typically do not consider visual features from images.

Contextual text block (CTB) detection is another approach for layout analysis that detects various text units, such as characters, words, or phrases , in reading order. 
CUTE~\cite{xue2022contextual} detects these text units and subsequently models relationships to cluster them (belonging to the same CTB) into an ordered sequence.
DRFormer~\cite{wang2024dynamic} uses a transformer decoder to generate a graph representing text relationships and dynamically refines the graph structure for CTB detection.

Hierarchical text detection methods implicitly link words. 
Unified Detector~\cite{long2022towards} detects scene text and forms text clusters.
Further, HTS~\cite{long2024hierarchical} uses a Unified-Detector-Polygon to detect text lines as Bézier curves and group lines into paragraphs; it uses a Line-to-Character-to-Word Recognizer to split lines into characters before merging them into words for hierarchical text spotting. 
LayoutFormer~\cite{liang2024layoutformer} uses three Transformer-based decoders to detect word-level, line-level, and paragraph-level text and adaptively learns word-line and line-paragraph relationships.
Text Grouping Adapter~\cite{bi2024text} leverages pretrained text detectors for layout analysis. 
While these methods effectively model hierarchical structures, they prioritize grouping text based on spatial proximity rather than explicitly linking words in reading order.

In comparison to these methods, \modelname~integrates Linguistic, Image/visual, and Geometric information, representing a text instance from multiple perspectives for linking Hierarchical Text on historical maps, thus allowing it to handle diverse and complex layouts.

\section{\modelname}  
\begin{figure}
    \centering
    \includegraphics[width=\textwidth]{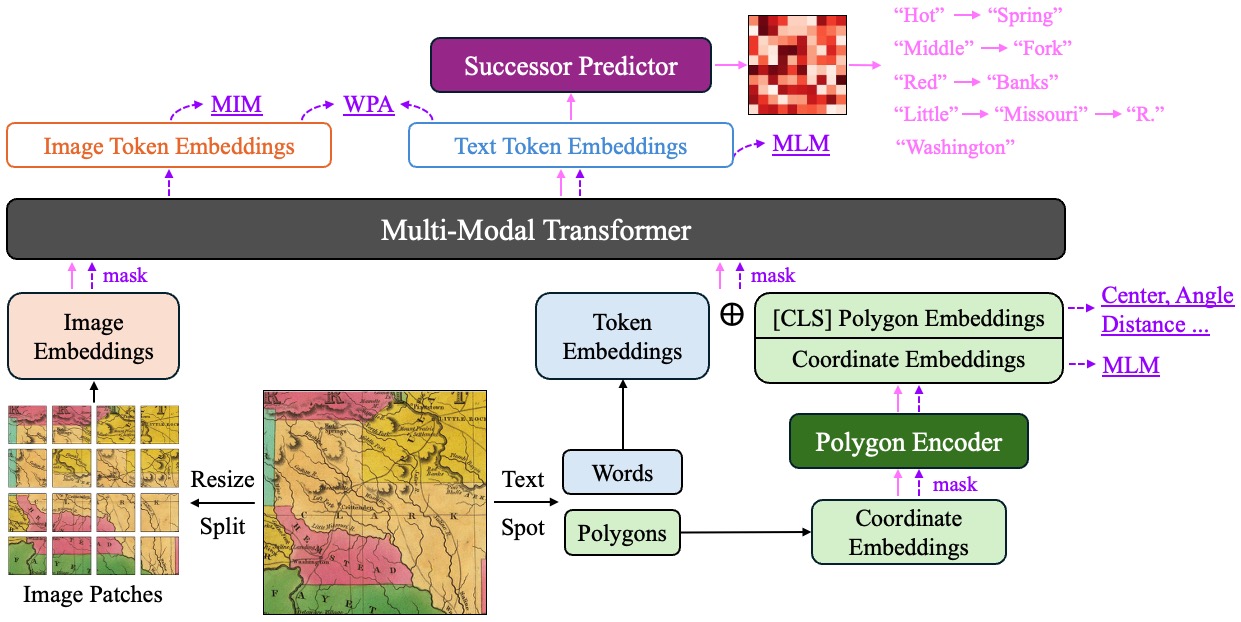}
    \caption{\modelname~contains three components: a Polygon Encoder, a Multi-Modal Transformer, and a Successor Predictor. Dashed purple arrows indicate the pretraining flow. Underlined purple labels are pretraining objectives for the two encoders. Solid pink arrows show the finetuning and inference flows.}
    \label{fig:model_architecture}
\end{figure}

Figure~\ref{fig:model_architecture} shows the overall architecture of \modelname~for linking text in reading order on historical maps.
Given a map image with the detected and recognized text instances, \modelname's polygon encoder first transforms the polygonal coordinates of each text instance into a fixed-size embedding. 
Then \modelname~integrates the polygon embeddings with the linguistic token embeddings and image features as input to a multi-modal transformer to learn a cross-modal representation for each word.
To adapt to historical map data, we pretrain both encoders using text labels and map images.
\modelname~leverages the learned representations to predict the successor of each word and construct text links.

\subsection{Polygon Encoder}
We propose a polygon encoder to capture the geometric characteristics of text polygons.
The polygon encoder builds on the BERT~\cite{devlin2019bert} architecture to encode an arbitrary-length sequence of coordinate points into a fixed-size embedding for each text instance.
After normalizing the coordinates $\left(X_i,Y_i\right)\in\mathbb{R}^2$ of a detected text instance to the range $\left[0,1\right]$ based on the input image size, we organize them into a sequence,
\[
[\texttt{CLS}], X_1, Y_1, X_2, Y_2, ... , X_m, Y_m, [\texttt{PAD}], [\texttt{PAD}], ...
\]
where $m$ is the number of control points defining the text polygon, which varies from word to word. 
We use the \texttt{[CLS]} token as the summary of the polygon and the \texttt{[PAD]} token to pad the sequence to a fixed maximum sequence length of 32.
Each coordinate is projected to a target embedding dimension ($D=768$) using a learnable linear layer. 
The [\texttt{CLS}] and [\texttt{PAD}] tokens are embedded using separate learnable embedding layers mapped to the same dimension. The sequence of embeddings is then fed into a BERT encoder.

One of the pretraining objectives for the polygon encoder is inspired by the masked language modeling (MLM) task in BERT. For each coordinate point~\((X_i, Y_i)\), we randomly mask either the \(X_i\) or \(Y_i\) dimension with a probability of~15\%. The model is then trained to reconstruct the masked values by minimizing the mean square error between predicted and ground truth coordinates. This objective, denoted as \(\mathcal{L_\text{MLM}}\), encourages the model to capture the relationships between coordinates and the geometric structure of a polygon. 


In addition to the MLM task, we introduce four auxiliary pretraining tasks on the [\texttt{CLS}] token to explicitly encode several polygon properties.
The [\texttt{CLS}] token embedding is trained to predict 
\begin{inparaenum}[1)]
\item the angle of the minimum enclosing rectangle for the polygon, \label{poly-aux-angle}
\item the center of polygon bounding box \((X_c, Y_c)\), \label{poly-aux-center}
\item the distance between the first and last points of the polygon (a proxy for text height assuming the polygon points start from the beginning of the text), and \label{poly-aux-dist}
\item the spatially closest other polygon.\label{poly-aux-closest} 
\end{inparaenum}
For the auxiliary tasks~\ref{poly-aux-angle}--\ref{poly-aux-dist}, we use multi-layer perceptrons (MLPs) to map the [\texttt{CLS}] token embedding to the desired output and use mean square error loss for regression training.
For task~\ref{poly-aux-closest}, we apply the dot product between all pair-wise [\texttt{CLS}] token embeddings, calculate the softmax, and predict the polygon index with the highest probability as the closest neighbor, guided by the Cross-Entropy loss.
The total loss is defined as:
\begin{equation}
\mathcal{L}_{\text{total}} = \mathcal{L_\text{MLM}}\ + \lambda_1 \mathcal{L}_{\text{angle}} + \lambda_2 \mathcal{L}_{\text{center}} + \lambda_3 \mathcal{L}_{\text{distance}} + \lambda_4 \mathcal{L}_{\text{closest}}
\label{eq:polygon-total-loss}
\end{equation}
where \(\lambda_1, \lambda_2, \lambda_3, \) and \(\lambda_4\) are weighting coefficients for the auxiliary tasks (all set as 0.1 in the experiment).
This multi-task pretraining strategy encourages the polygon encoder to effectively learn the local and global geometric properties of text polygons within an image.
The [\texttt{CLS}] token embedding serves as the embedding to summarize the polygon, denoted as $E_{\text{poly}}$.

\subsection{Multi-Modal Transformer}
\modelname~employs a multi-modal transformer to generate cross-modal representations for text instances. 
Existing document analysis methods, such as LayoutLMv3~\cite{huang2022layoutlmv3}, jointly model text and image but overlook text shapes (or only use bounding boxes to indicate text positions), which is crucial for handling irregularly shaped text on historical maps.
\modelname~builds on LayoutLMv3 by incorporating polygon embeddings, i.e., [\texttt{CLS}] token embeddings, generated by the polygon encoder to enrich geometric information. 

First, an embedding layer transforms the tokenized word into text token embeddings.
As shown in Figure~\ref{fig:model_architecture}, \modelname~adds the polygon embeddings to the text token embeddings, followed by layer normalization and dropout.
Text tokens belonging to the same word share the same polygon embedding of the word.
The polygon embeddings serve as a polygon summarization method while preserving polygon properties (e.g., angle, position, height, and relative distance), which provides detailed geometric information for text layout understanding on maps.
Therefore, this encoder takes a sequence of text token embeddings, polygon embeddings, and image features as input to a multi-modal transformer, fusing linguistic, visual, and geometric information layer by layer. 


We pretrain the multi-modal transformer with the polygon encoder on map text and images.
To obtain text annotations for pretraining, instead of using the OCR tool in LayoutLMv3, we use a text spotting model specialized in historical maps~\cite{lin2024hyper} to detect and recognize text instances on map images, which generates a large amount of training data (details in Section~\ref{sub:pretraining-data}).
We follow the pretraining steps outlined in LayoutLMv3, utilizing three objectives: masked language modeling (MLM), masked image modeling (MIM), and word-patch alignment (WPA).
For MLM, we randomly mask 30\% of the text tokens, and the pretraining objective is to reconstruct the masked tokens based on the contextual representations of the corrupted sequence of text and image tokens and geometric information added to the text tokens.
For MIM, we randomly mask 40\% of the image tokens, and similarly, the objective is to reconstruct the masked image tokens~\cite{bao2022beit}.
We use a pretrained image tokenizer to generate the labels of image tokens, which can transform dense image pixels into a discrete token with a visual vocabulary~\cite{ramesh2021zero}.
For WPA, the intuition is to perform explicit alignment learning between text and image modalities, as the masking processes are independent for MLM and MIM.
The objective is to predict whether the corresponding image patches of a text word are masked. By training with these three objectives on map data, the model can capture cross-modal relationships between map text (contents and shapes) and images.
\modelname~uses the final token embeddings for the text linking task, denoted as $E_{\text{text}} \in \mathbb{R}^{N\times D}$, where $N$ is the number of words and $D$ is the embedding dimension ($D=768$). 


\subsection{Link Prediction and Bidirectional Learning Strategy}

\subsubsection{Successor Prediction}
The link prediction task starts from a text instance and considers all other instances within a map image as candidate successors to generate a text sequence.
If \modelname~predicts the successor of a text instance is itself, that instance is marked as the end of the linking path.
Conversely, if the model predicts another instance as the successor of a text instance, the identified instance is linked as next in the sequence.
Note that at this step, \modelname~predicts only the ``forward'' direction for each text instance.
This task uses the first token embedding of each word to represent the entire word for link prediction, inspired by sequence labeling tasks such as named entity recognition~\cite{devlin2019bert}.
To model directional relationships, we apply two dimension-preserving MLPs---each consisting of a Linear-ReLU-Linear architecture---to project the final token embeddings into two separate embeddings, representing the words as predecessors and successors, respectively. 
\begin{equation}
E_{\text{pre}} = \text{MLP}_{\text{pre}}(E_{\text{text}}), \quad E_{\text{succ}} = \text{MLP}_{\text{succ}}(E_{\text{text}})
\label{eq:link-pred-succ}    
\end{equation}
Then we compute an $N\times N$ association score matrix \( S \) as the matrix product of the two embedding matrices
\begin{equation}
S = E_{\text{pre}} \cdot E_{\text{succ}}^\top
\label{eq:link-pre-succ-matrix}
\end{equation}
Each element  \( s_{ij} \) of \( S \) is the logit for the probability of a directional link from word $i$ to word $j$ (so $j$ is a successor of $i$).
We denote the probability matrix as~$P \in \mathbb{R}^{N\times N}$.
The probability $p_{ij}$ in $P$ arises from computing a softmax over row $i$ of $S$ as,
\begin{equation}
p_{ij} = \frac{\exp(s_{ij})}{\sum_{k} \exp(s_{ik})}.
\end{equation}

\subsubsection{Bidirectional Learning and Optimization}
We compute the Cross-Entropy loss over the predicted probability matrix $P$ and the ground truth text links in a map image, denoted as~$\mathcal{L}_{\text{CE}}$. 

However, a significant imbalance exists: 
(1) Most word pairs do not have links in between, resulting in a highly sparse ground truth matrix; 
(2) The number of words with self-successors is significantly greater than those without, since we label the end of the linking path as the word itself.
This imbalance causes the model to overly predict ``no link'' or ``self links'', reducing its ability to learn meaningful text relationships.
Therefore, in addition to the Cross-Entropy loss, we adopt a focal loss~\cite{lin2017focal} to encourage the model to focus on hard-to-predict linkages, and also assign lower weights to self-successor links to prioritize learning links between words:
\begin{equation}
\mathcal{L}_{\text{focal}} = -\sum_{i} \alpha_i \left( y_i (1 - p_i)^\gamma \log p_i + (1 - y_i) p_i^\gamma \log (1 - p_i) \right).
\label{eq:focal-loss}
\end{equation}
where $p_i$ is the predicted probability of a link in $P$, 
$y_i$ is the target label where~1 represents a successor relationship and 0 is for a non-successor,
$\gamma$ is the focusing parameter (set as 2), and
$\alpha_i$ is the balancing factor (set as 0.25 for self-successor links and 1 for all others).

To further enhance the model’s capability to learn the relational structure between words, \modelname~incorporates a bidirectional linking mechanism that simultaneously considers both successor and predecessor relationships. \modelname~therefore computes a \emph{reverse} score matrix (with a corresponding softmax $P'$):
\begin{equation}
S' = E_{\text{succ}} \cdot E_{\text{pre}}^\top.
\label{eq:link-succ-pre-matrix}
\end{equation}
The bidirectional prediction strategy enables \modelname~to learn both forward and backward linkages between words, providing a robust prediction of the text order.
Similar to the forward direction, we apply the Cross-Entropy loss on $P'$ to maximize the log-likelihood of the predecessors ($\mathcal{L}^\text{bi}_{\text{CE}}$) and a focal loss to mitigate the imbalance issue ($\mathcal{L}^\text{bi}_{\text{focal}}$).
The final training objective is formulated as
\begin{equation}
\mathcal{L} = \mathcal{L}_{\text{CE}} + \mathcal{L}_{\text{focal}} + \mathcal{L}^\text{bi}_{\text{CE}} + \mathcal{L}^\text{bi}_{\text{focal}}.
\label{eq:overall_loss}
\end{equation}

\subsubsection{Link Inference} During inference, \modelname~constructs text linking paths by iteratively selecting the most probable successor for each word in a greedy manner.
Given a target word, \modelname~identifies the next word with the highest predicted probability as its successor.
However, this process can lead to ambiguities where a word is assigned to multiple predecessors.
To ensure a word belongs to only one phrase, \modelname~enforces a selection rule: if a word has multiple predecessors, only the word with the highest probability is retained as the unique predecessor, while other predecessors continue searching for their next-best successor. 
This greedy strategy ensures that each word is linked to at most one predecessor, preventing conflicts and preserving a clear, coherent reading order for text. Then \modelname~generates sequences from these word-to-successor links. We provide the pseudo-code in the supplementary material Section~\ref{app: alg}.

\section{Experiments}
\subsection{Datasets}
\subsubsection{Text Linking Dataset} We use the Rumsey benchmark dataset~\cite{lin_2024_11516933,lin_2024_10776183} from the ICDAR 2024/2025 MapText Competitions~\cite{li24icdar,lin25icdar} for model training (200 tiles), validation (40 tiles), and testing (700 tiles). Each map tile has a resolution of $2{,}000 \times 2{,}000$ pixels. We provide the dataset statistics in the supplementary material Section~\ref{app: dataset}.

\subsubsection{Pretraining Dataset}\label{sub:pretraining-data} We leverage additional map images from the David Rumsey Map Collection~\cite{RumseyCollection}~to pretrain \modelname's polygon encoder and multi-modal transformer using self-supervised learning.
We generate 397,385 cropped map tiles from 41,279 maps with a mix of both $1{,}000 \times 1{,}000$ and $2{,}000 \times 2{,}000$ pixels.
To produce word-level detections and transcriptions, we run a state-of-the-art map text spotting model~\cite{lin2024hyper}, which was pretrained on synthetic map image and  scene image datasets, and finetuned on the MapText Rumsey training data.
These automatically generated polygons and text transcriptions serve as labels for pretraining.

\subsection{Experimental Setup}
\subsubsection{Tokenization}
Given a map image and text labels, \modelname~gathers all the individual words together into a ``sentence.'' 
During training, the words are shuffled to avoid bias toward the ground truth ordering.
During testing and inference, words are sorted top-to-bottom (by $Y$ coordinate) and then left-to-right (by $X$ coordinate).
For tokenization, we use the pretrained LayoutLMv3 tokenizer, based on RoBERTa with Byte Pair Encoding~\cite{liu2019roberta}.
This tokenizer processes the input words along with their bounding boxes, polygons, and group labels, generating token-level inputs---\texttt{input\_ids}, \texttt{attention\_mask}, \texttt{bounding\_boxes}, \texttt{polygons}, and \texttt{labels}.
We set the maximum sequence length to 1,000.

To limit computational complexity, we preprocess map images by resizing the entire input tile to a fixed size of $224\times224$ pixels.
During pretraining, we generate image patches and utilize the BEiT image tokenizer~\cite{bao2022beit} to transform patches into image token ids as the ground truth.
During training (finetuning), we pass the resized image into \modelname~to generate visual features.

\subsubsection{Training Settings}
For the polygon encoder, we set the maximum sequence length as 32, the number of layers as 6, and the embedding dimension as 768.
Pretraining is conducted for 100,000 steps with a batch size of 8, using a maximum learning rate (LR) of 0.0001, a 10\% warm-up phase, and a linear LR decay~\cite{koroteev2021bert}.
The multi-modal transformer adopts 12 layers and a 768-dimensional embedding size, consistent with LayoutLMv3$_\mathrm{BASE}$.
Pretraining runs for 375,000 steps with a batch size of 4, using a maximum LR of 0.0001, a 4.8\% warm-up phase, and linear decay~\cite{huang2022layoutlmv3}.
During finetuning, we use a batch size of 2 and an initial LR of 0.0005, with a performance-based LR scheduler.
If validation performance does not improve for 5 epochs, the LR is reduced by 10\%; 
training is terminated after 9 consecutive non-improving epochs. All the models are trained with one NVIDIA A100 (40GB) GPU.\footnote{Code and data:  \url{https://github.com/knowledge-computing/multimodal-text-linking}}

\subsection{Evaluation Metric}
We adopt the evaluation metrics from the ICDAR 2025 MapText Competition~\cite{maptext25,lin25icdar}.\footnote{Repo: \url{https://github.com/icdar-maptext/evaluation}}
Previous benchmarks, such as HierText~\cite{long2022towards,long2023icdar}, use group Panoptic Detection Quality (PDQ) metrics for group-level evaluation. 
However, group PDQ tends to penalize entire phrases for missing a single link rather than assessing individual link accuracy.
Instead, we follow the 2025 competition and report link-level precision $P_L$, recall $R_L$, and F-score $F_L$, which directly evaluate the correctness of predicted links between words.
We also report the harmonic mean (H-Mean) $\overline{H}$ of all quantities for the end-to-end competition task of text detection, segmentation, recognition, and phrase linking.

\subsection{Results and Discussion}
This section presents results and analysis in four parts:
\begin{inparaenum}[1)]
\item a comparison between \modelname~and two non-linguistic text linking baselines, \label{exp1}
\item an ablation study on individual input modalities---language (L), image (I), and geometry (G), \label{exp2}
\item an analysis of the impact of the proposed learning objectives, and \label{exp3}
\item a comparison with other submissions from the ICDAR 2024/2025 MapText competitions.\label{exp4} 
\end{inparaenum}
The first three analyses assume perfect (ground truth) text detection and recognition, while the last uses actual (imperfect) text spotter output.
All models are trained under identical conditions.

\subsubsection{Comparison with Non-Linguistic Baselines}\label{sub:exp1} 
Table~\ref{tab: non-ling} compares the performance of \modelname~with two non-linguistic methods: (1) character distance threshold (grouping words with distance $<$ 2 characters)~\cite{kim2023mapkurator}, and (2) heuristic MST~\cite{olson2024automatic}.
The baseline methods might generate more than two bi-directional links for a word, whereas \modelname~predicts directional links with at most one successor and one predecessor.
For a fair comparison, we apply an edge-cutting step to the baselines by iteratively removing the highest-weighted edge (based on the edge cost heuristic from Olson et al.~\cite{olson2024automatic}) until all words have at most two links. We then sort the words in each path by $Y$-coordinate and then by $X$-coordinate to form an ordered sequence (i.e., phrase). 
The results show that non-linguistic methods achieve high recall but significantly lower precision compared to \modelname, indicating their tendency to over-predict links.
Because these methods do not consider any linguistic context, words with similar shapes or small distances are grouped, which might suffice in the multi-word search setting (see Section~\ref{related: 2.1}). 
\begin{table}[h]
    \centering  
    \vspace{\tabvspace}
    \caption{Comparison between \modelname~and non-linguistic baselines.}
    \setlength{\tabcolsep}{5pt} 
    \begin{tabular}{lccc} \toprule
        Method & $R_L$ & $P_L$ & $F_L$ \\ \midrule
        Character Distance Threshold \cite{kim2023mapkurator} & 62.6 & 43.6 & 51.4 \\
        Heuristic MST~\cite{olson2024automatic} & 62.5 & 34.3 & 44.3 \\ \cmidrule{1-4}
        \modelname                              & \textbf{81.2} & \textbf{86.3} & \textbf{83.7} \\ \bottomrule
    \end{tabular}
    \vspace{\tabvspace}
    \label{tab: non-ling}
\end{table}

\subsubsection{Ablation Study on Input Modalities} 
Table~\ref{tab: modalities} compares models using different combinations of input modalities: Language, Image, and Geometry.
All models are finetuned with the same learning objectives and prediction head as \modelname.
First, \modelname's polygon encoder, which uses only geometry information, outperforms the non-linguistic baselines in Table~\ref{tab: non-ling} but remains the worst among the variants. 
BERT~\cite{devlin2019bert} uses only linguistic features, achieving the same $F_L$ score with higher precision but lower recall than \modelname's polygon encoder.
This result suggests that textual content can provide useful information for linking, but has limitations in identifying complete links.
Adding visual features via LayoutLMv3\footnote{We finetuned the model on map data from the pretrained weights on Hugging Face.}~\cite{huang2022layoutlmv3} boosts the performance over the language-only model (BERT), demonstrating the benefits of incorporating visual cues.
Furthermore, we re-pretrain LayoutLMv3 on our map pretraining dataset (Section~\ref{sub:pretraining-data}), followed by finetuning as \modelname.
This domain-adaptive pretraining leads to a substantial 13.2\% gain in F-score.
Finally, \modelname~integrates information from all three modalities, i.e., language, image, and geometry, achieving the best performance with a 5.9\% improvement in F-score.
This result highlights the complementary nature of the three modalities in text linking for historical maps. 
\begin{table}[h]
    \centering
    \vspace{\tabvspace}
    \caption{Ablation study on modalities: Language (L), Image (I), Geometry (G)}
    \setlength{\tabcolsep}{5pt} 
    \begin{tabular}{lcccc} \toprule
        Method & Modality & $R_L$ & $P_L$ & $F_L$ \\ \midrule
        LIGHT - Polygon Encoder & (G) & 57.3 & 62.5 & 59.8 \\ 
        BERT$_\mathrm{BASE}$~\cite{devlin2019bert} (finetune) & (L)  & 52.2 & 70.3 & 59.8 \\
        LayoutLMv3$_\mathrm{BASE}$~\cite{huang2022layoutlmv3}  (finetune) & (L)(I) & 54.4 & 73.8 & 64.6 \\ 
        LayoutLMv3$_\mathrm{BASE}$ (pretrain+finetune) & (L)(I) & 73.5 & 82.6 & 77.8 \\ \cmidrule{1-5}
        \modelname                     & (L)(I)(G) & \textbf{81.2} & \textbf{86.3} & \textbf{83.7} \\ \bottomrule
    \end{tabular}
    \vspace{\tabvspace}
    \label{tab: modalities}
\end{table}


\subsubsection{Ablation Study on Learning Objectives} 
We further examine the impact of the bidirectional learning strategy and focal loss. We observe that without auxiliary losses (\modelname-plain), the model achieves an F-score of 74.2\% and successfully identifies about 70\% of links.
Introducing focal loss (\modelname-focal) significantly enhances performance, leading to a 4.6\% improvement in F-score.
Encouraging the model to focus on hard-to-predict and non-self links is beneficial given the high number of words without successors. 
Moreover, incorporating the bidirectional loss (\modelname-bidirectional) also shows a 3.4\% performance boost over \modelname-plain.
Combining these losses results in the best performance in \modelname, demonstrating the complementary benefits of focusing on challenging links and leveraging bidirectional relationships.

\begin{table}
    \centering
    \vspace{\tabvspace} 
    \caption{Ablation study on learning objectives in \modelname.}
    \setlength{\tabcolsep}{5pt} 
    \begin{tabular}{lccccccc} \toprule
        Method                         & $\mathcal{L}_{\text{CE}}$ & $\mathcal{L}_{\text{focal}}$ & $\mathcal{L}^\text{bi}_{\text{CE}}$ & $\mathcal{L}^\text{bi}_{\text{focal}}$ & $R_L$ & $P_L$ & $F_L$ \\ \midrule
        \modelname-plain               & \checkmark                &                              &                                     &                                        & 70.4  & 78.4  & 74.2 \\
        \modelname-focal               & \checkmark                & \checkmark                   &                                     &                                        & 74.7  & 83.3  & 78.8 \\ 
        \modelname-bidirectional       & \checkmark                &                              & \checkmark                          &                                        & 74.1  & 81.5  & 77.6 \\ \cmidrule{1-8}
        \modelname (full)              & \checkmark                & \checkmark                   & \checkmark                          & \checkmark                             & \textbf{81.2} & \textbf{86.3} & \textbf{83.7} \\ \bottomrule
    \end{tabular}
    \label{tab: overall}
\end{table}

\subsubsection{Comparison with Competition Submissions}
We further test \modelname~with actual (imperfect) text and polygon predictions from text spotting models.
We compare \modelname~to the top-performing submissions in the ICDAR 2024 and 2025 MapText Competitions~\cite{li24icdar,lin25icdar}.\footnote{Although placing third by the 2024 metric, DS-LP is the only competitor to produce any links at all---an artifact of the original competition's PQ-based metric.} For MapText Strong and Self-Sequencer, whose polygon outputs contain 50 points, we uniformly resample them to 16 clockwise points to meet the input requirements of \modelname’s polygon encoder. This adjustment preserves or slightly improves their word-level end-to-end performance due to the correction of invalid polygons. Integrated with these spotting models, \modelname~consistently outperforms their original linking modules, i.e., achieving a 19.9\% F-score gain in DS-LP and 7.8\% in Self-Sequencer. Although MapText Strong lacks a linking component, its high spotting quality enables strong linking performance when combined with \modelname.
We also test \modelname~with spotting outputs from P\textsc{a}L\textsc{e}TT\textsc{e}, a state-of-the-art text spotter for historical maps~\cite{lin2024hyper}. We observe that higher spotting precision typically leads to improved linking performance. Overall, \modelname~shows robust link prediction even with imperfect spotting labels.
\begin{table}
    \setlength{\tabcolsep}{5pt}
    \centering
    \vspace{\tabvspace}
    \caption{Comparison applying \modelname~($\oplus$) to baseline methods' text spotting (detection and recognition) output.  $T$ is tightness (average IoU), $C$ is the character accuracy, and $\overline H$ (H-Mean) is the overall competition ranking metric.}
    \resizebox{\textwidth}{!}{
    \begin{tabular}
{lccccccccc}
    \toprule
         \multicolumn{1}{c}{\multirow{2}{*}{Method}} & \multicolumn{3}{c}{Links} & \multicolumn{5}{c}{Words} & \multicolumn{1}{c}{\multirow{2}{*}{$\overline H$}} \\ \cmidrule(lr){2-4}  \cmidrule(lr){5-9} 
 & $R_L$ & $P_L$ & $F_L$ & $R$ & $P$ & $F$ & $T$ & $C$ \\ \midrule
        DS-LP~\cite{li24icdar} & 16.6 & 55.3 & 25.6 & 78.9 & 71.8 & 75.2 & 71.6 & 90.8 & 46.2\\ 
        $\quad \oplus$\modelname & 41.4 & 50.5 & 45.5 & " & " & " & " & " & 62.8 \\ \cmidrule(lr){1-10}

        MapText Strong~\cite{lin25icdar} & {-} & {-} & {-}& 91.8 & 95.9 & 93.8 & 83.8 & 94.0 & {-} \\
        $\quad \oplus$\modelname & 68.5 & \textbf{81.6} & \textbf{74.5} & 91.8 & 95.8 & 93.7 & 82.6 & 94.1 & 84.6 \\ \cmidrule(lr){1-10}        

        Self-Sequencer~\cite{zou2025recognizing}  & 61.7 & 72.6 & 66.7 & 89.1 & 91.5 & 90.3 & 86.1 & 94.9 & 80.8 \\ 
        $\quad \oplus$\modelname & \textbf{69.1} & 80.9 & \textbf{74.5} & 91.0 & 93.7 & 92.3 & 86.0 & 94.7 & \textbf{84.9} \\ \cmidrule(lr){1-10}


        \modelname/P\textsc{a}L\textsc{e}TT\textsc{e}~\cite{lin2024hyper} & 68.5 & 77.1 & 72.6 & 92.2 & 88.9 & 90.5 & 85.8 & 94.6 & 83.5 \\ \bottomrule
    \end{tabular}}
    \vspace{\tabvspace}
    \label{tab: overall2}
\end{table}

\subsubsection{Visualizations} 
\begin{figure}[h]
    \centering
    \begin{subfigure}[b]{0.49\textwidth}
        \centering
        \fbox{\includegraphics[width=\linewidth]{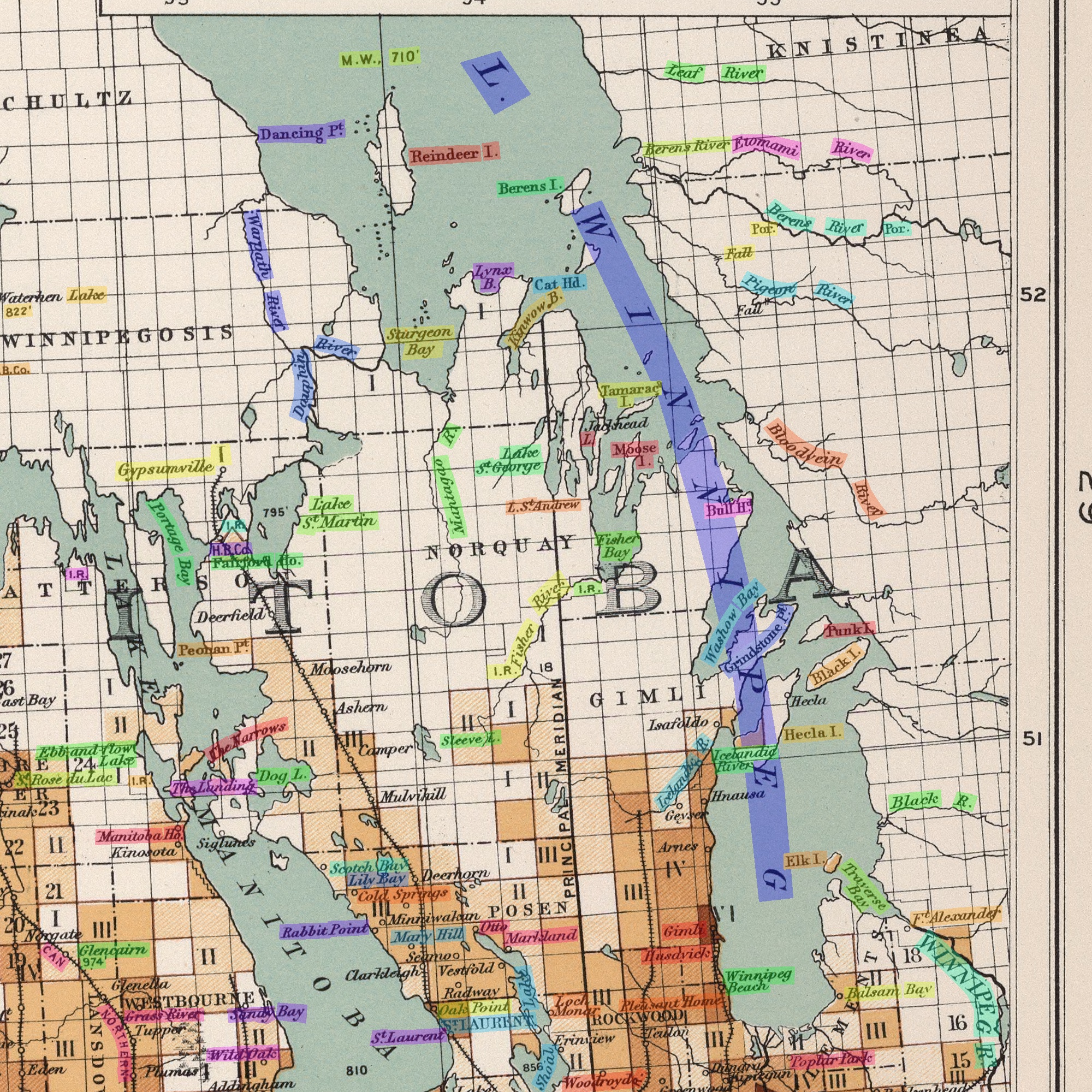}}
    \end{subfigure}
    \begin{subfigure}[b]{0.49\textwidth}
        \centering
        \fbox{\includegraphics[width=\linewidth]{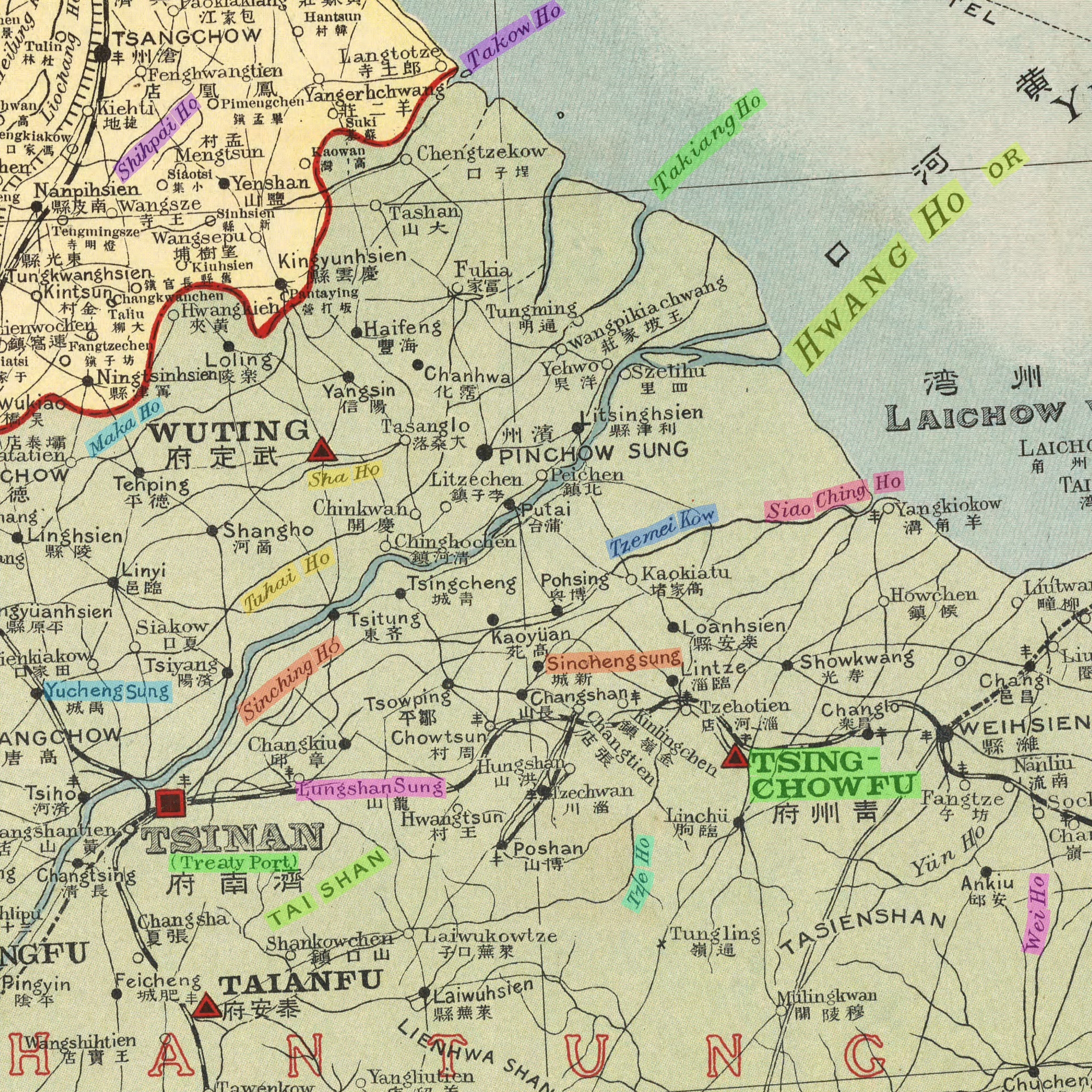}}
    \end{subfigure}
    
    \vspace{0.2cm} 

    \begin{subfigure}[b]{0.49\textwidth}
        \centering
        \fbox{\includegraphics[width=\linewidth]{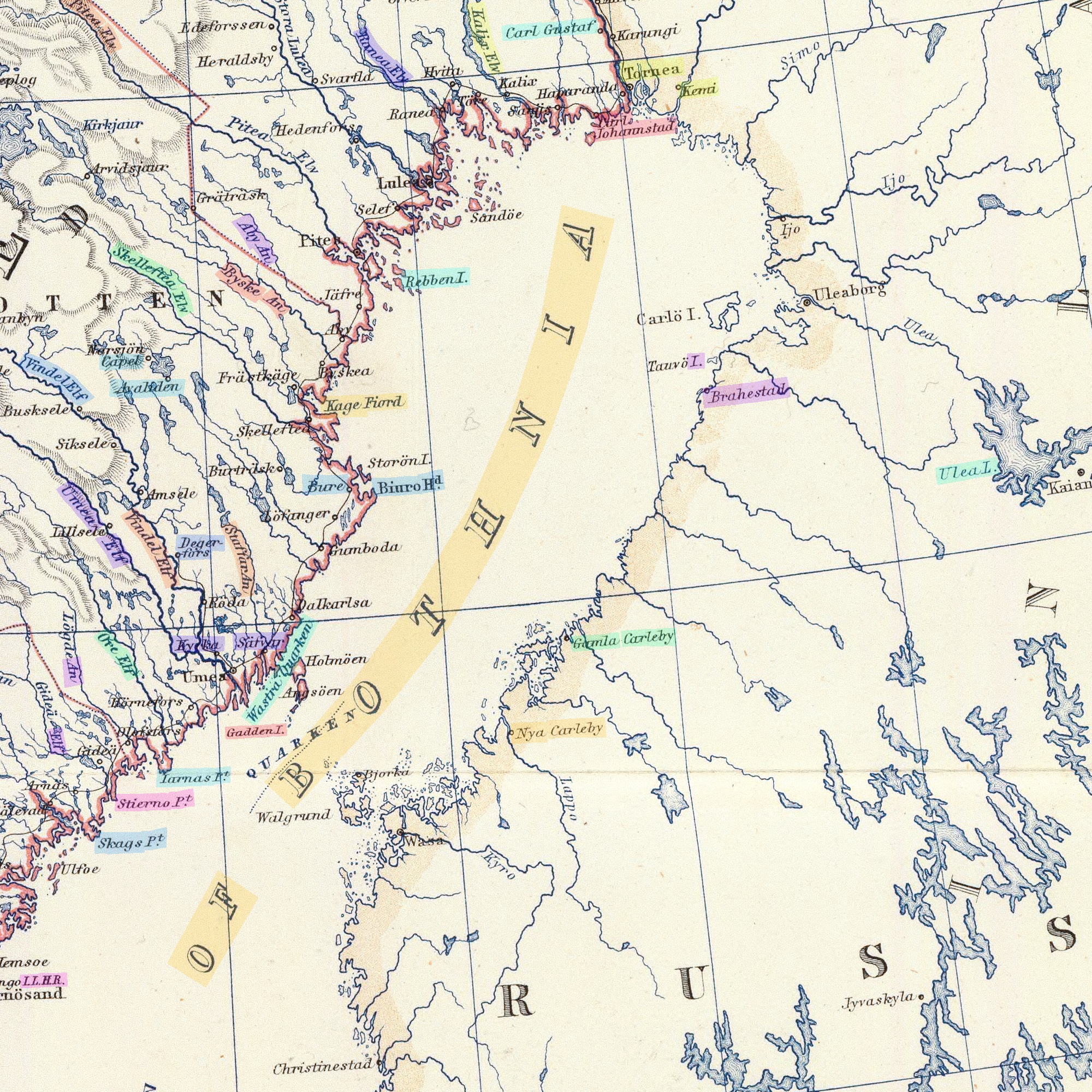}}
    \end{subfigure}
    \begin{subfigure}[b]{0.49\textwidth}
        \centering
        \fbox{\includegraphics[width=\linewidth]{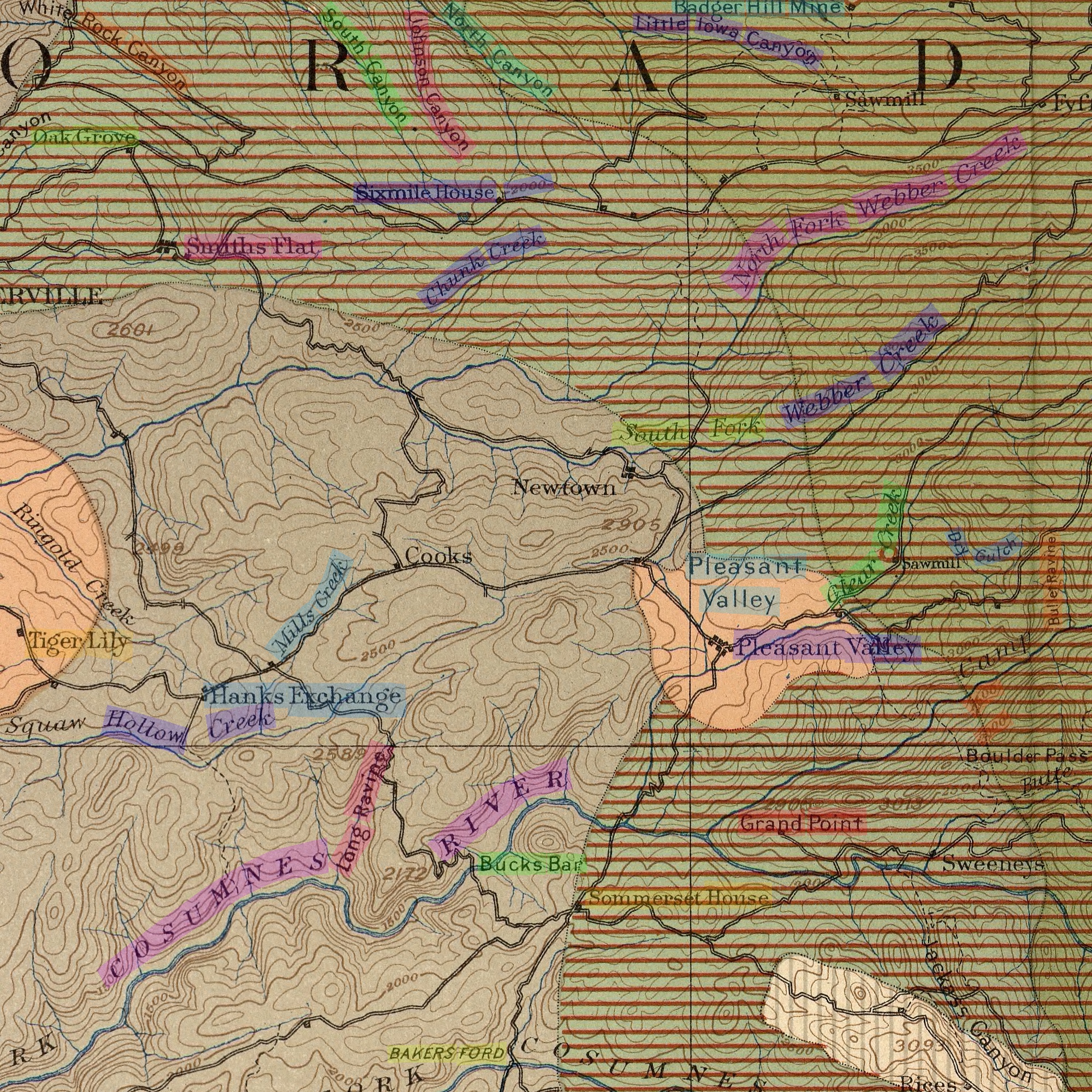}}
    \end{subfigure}
    \caption{Successful examples with link recall $\geq$90\%} 
    \label{fig: good}
\end{figure}

\begin{figure}[h] 
    \centering
    \begin{subfigure}[b]{0.49\textwidth}
        \centering
        \fbox{\includegraphics[width=\linewidth]{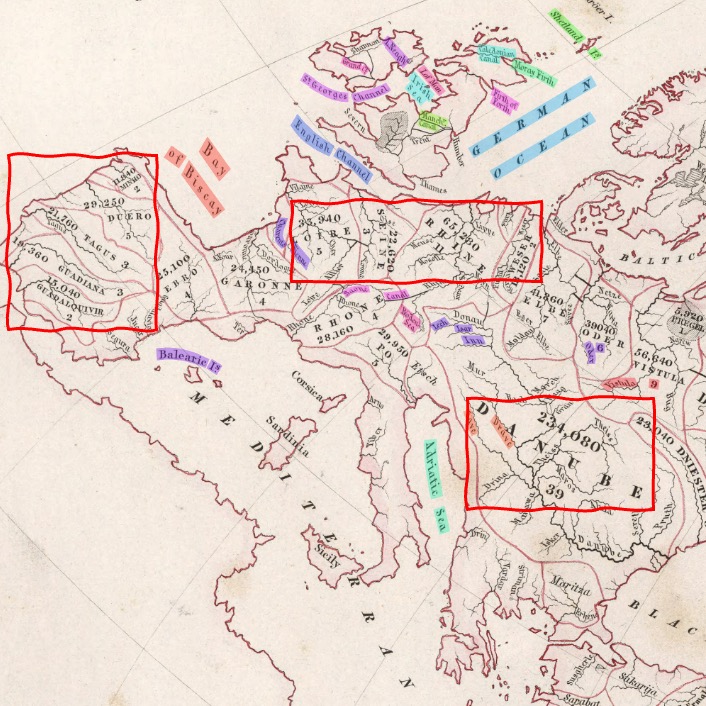}}
    \end{subfigure}
    \begin{subfigure}[b]{0.49\textwidth}
        \centering
        \fbox{\includegraphics[width=\linewidth]{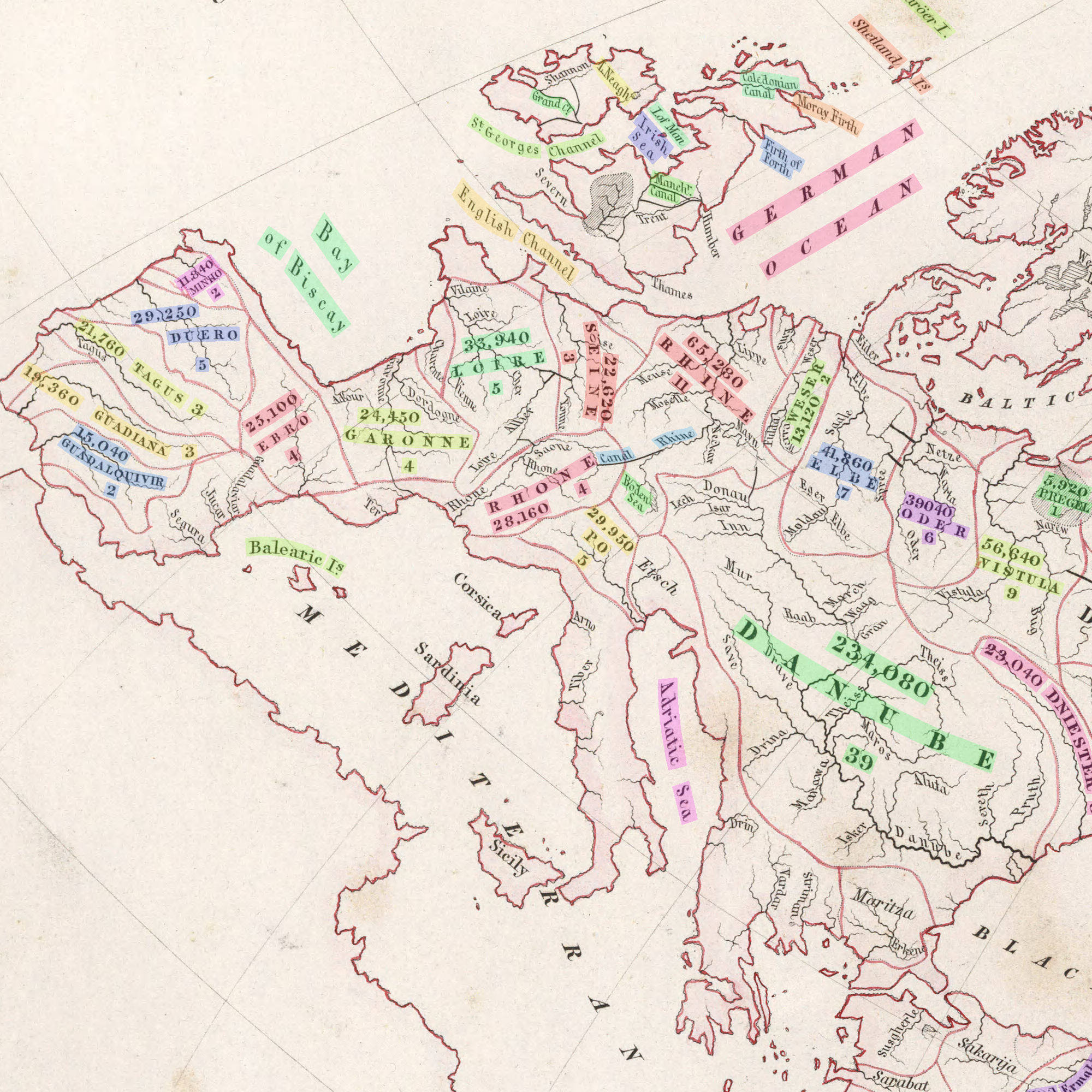}}
    \end{subfigure}

    \vspace{0.2cm} 

    \begin{subfigure}[b]{0.49\textwidth}
        \centering
        \fbox{\includegraphics[width=\linewidth]{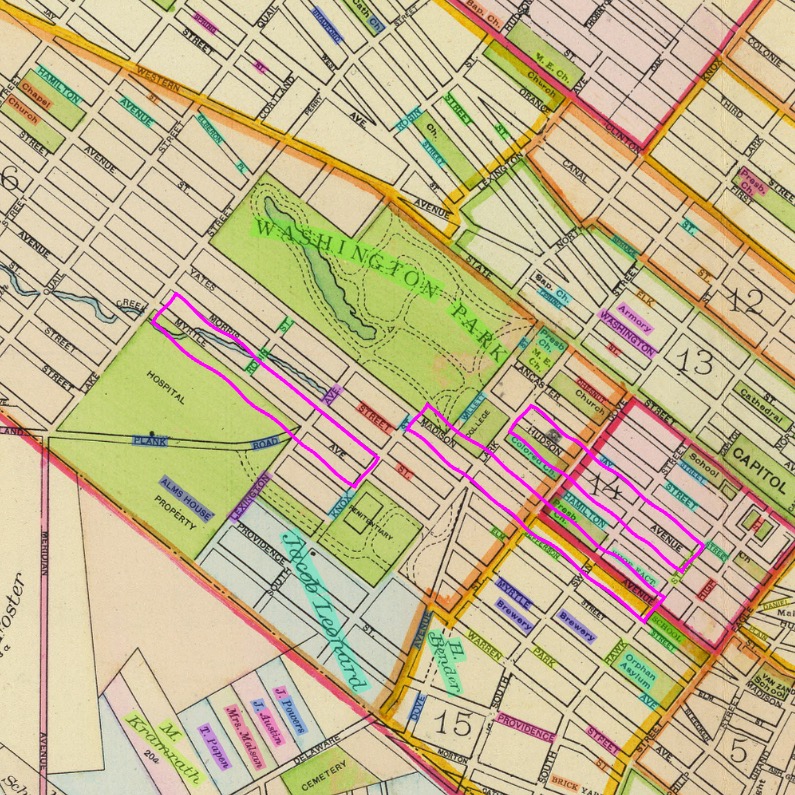}}
    \end{subfigure}
    \begin{subfigure}[b]{0.49\textwidth}
        \centering
        \fbox{\includegraphics[width=\linewidth]{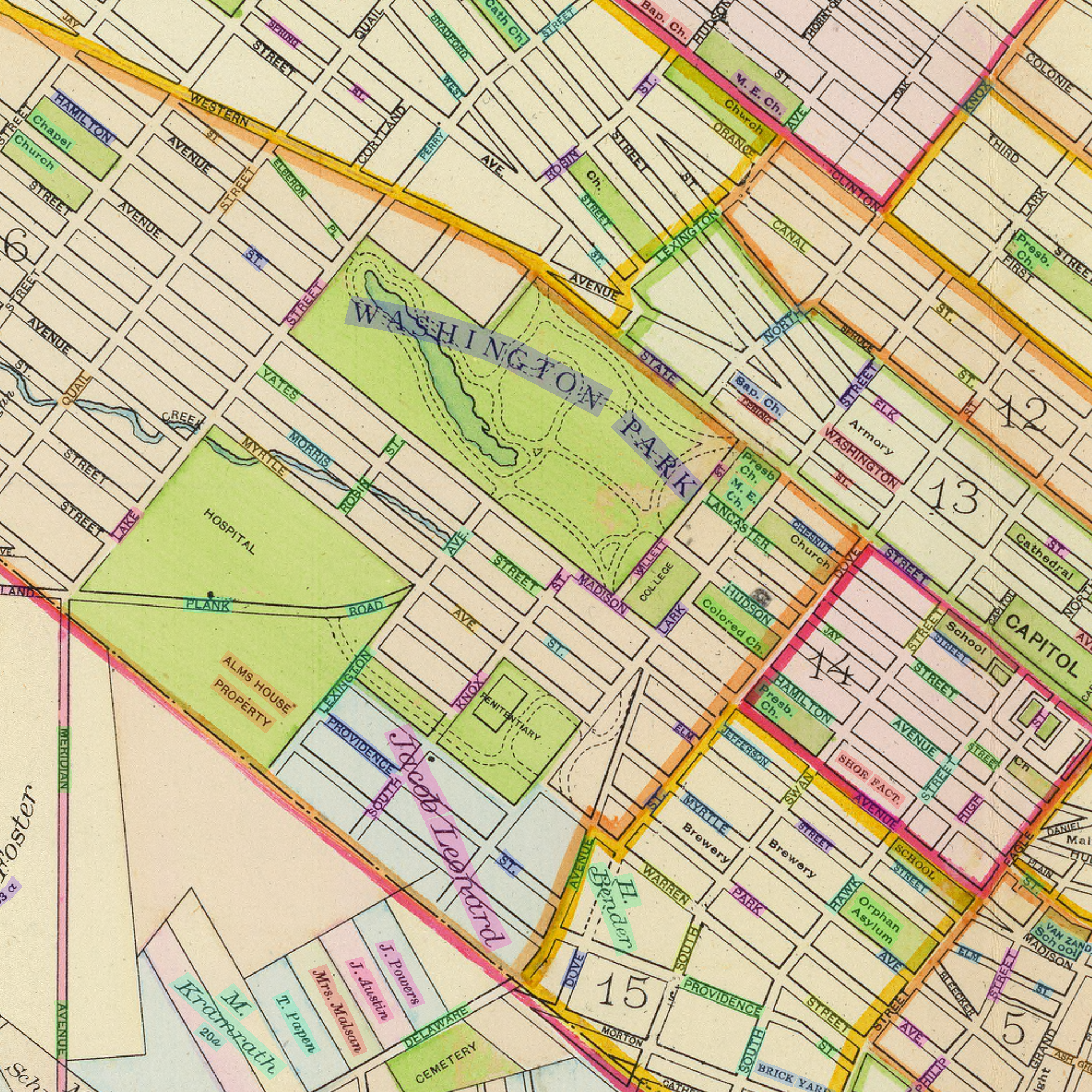}}
    \end{subfigure}
    \caption{Examples of poor linking results (left) and ground truth (right)}  
    \label{fig: poor}
\end{figure}

Over 400 of the 700 test map tiles have $\geq$80\% of links correctly identified. 
Figure~\ref{fig: good} presents three strong examples where \modelname~accurately links words in dense text regions and spatially far apart text, demonstrating its robustness to handle diverse layouts on historical maps. 
However, a few maps show relatively low performance. We show two examples in Figure~\ref{fig: poor}. In the first example, the model fails to group words mixed with numbers (e.g., red boxes). This group pattern may be rare or absent in the training set (the training set is much smaller than the testing set), making it challenging for the model to predict correct links. In the second example, the difficulty arises from some street or avenue names where words are positioned extremely far apart to indicate the entire roads (magenta boxes). While humans can intuitively infer these links by following road lines, the current model does not explicitly incorporate such information, leading to missed connections.



\setlength{\fboxrule}{0.1pt}   
\setlength{\fboxsep}{0pt}    

\section{Conclusion, Limitations, and Future Work}
We introduce \modelname, a novel multi-modal text linking approach on historical maps. 
\modelname~integrates three modalities, i.e., language, vision, and geometry, to predict the successor of a given word, thereby identifying link paths for text on map images.
\modelname~achieves state-of-the-art performance, outperforming all baselines, including leading methods from the ICDAR 2024/2025 MapText competitions.
The extracted phrases can support downstream tasks such as georeferencing and semantic typing.
Despite its strong performance, \modelname~is not a fully end-to-end approach (i.e., one that integrates text detection, recognition, and linking into a unified framework), hence it relies on external text spotting models and processes downsampled images (from 2000$\times$2000 to 224$\times$224), limiting its capability to learn localized visual features. 
For future work, \modelname~could be extended into a unified end-to-end approach from text detection, recognition, to linking. Such integration will enable the linking module to benefit from rich visual cues extracted by the detection and recognition modules, potentially improving the overall performance on diverse historical map styles and complex text layouts.

\begin{credits}
\subsubsection{\ackname} This material is based upon work supported in part by the National Science Foundation under Grant No. BCS-2419334. This work is also supported by the Doctoral Dissertation Fellowship, 2024--2025, from the University of Minnesota Graduate School. The authors also thank the David Rumsey Historical Map Collection and the generous support from David and Abby Rumsey.
\end{credits}

\bibliographystyle{splncs04}
\bibliography{ref}

\clearpage
\chapter*{Supplementary Material}
\appendix


\setcounter{figure}{0}
\renewcommand{\thefigure}{S\arabic{figure}}

\setcounter{table}{0}
\renewcommand{\thetable}{S\arabic{table}}

\setcounter{equation}{0}
\renewcommand{\theequation}{S\arabic{equation}}

\setcounter{algorithm}{0}
\renewcommand{\thealgorithm}{S\arabic{algorithm}}

\appendix
\section{Complexity Analysis}\label{app: complex}
Table~\ref{tab: comp} shows the number of parameters and inferencing time per image for the baselines and \modelname. While \modelname, which integrates a BERT-based polygon encoder and LayoutLMv3, has the highest computational cost, it also delivers significantly better performance than all baselines. Since the text linking task does not require real-time inference, the runtime remains well within acceptable limits for practical use.

\begin{table}[h]
    \centering
    \vspace{-0.2in}
    \caption{Number of parameters and inferencing time (second per image).}
        \setlength{\tabcolsep}{5pt} 
        \begin{tabular}{lSS} \toprule
            Method                     & {\#Param ($\times 10^6$)} & {Time (s/image)} \\ \midrule
            LIGHT - Polygon Encoder    & 70.1  & 0.16 \\ 
            BERT$_\mathrm{BASE}$       & 111.0 & 0.17 \\
            LayoutLMv3$_\mathrm{BASE}$ & 130.4 & 0.19\\ 
            \modelname                 & 197.1 & 0.34 \\ \bottomrule
        \end{tabular}
    \label{tab: comp}
    \vspace{-0.2in}
\end{table}

\section{Dataset Statistics}\label{app: dataset}
Table~\ref{tab: data stats} shows the statistics of the Rumsey benchmark dataset used for model training, validation, and testing in the experiment.
\begin{table}[h]
    \setlength{\tabcolsep}{11pt}
    \centering
    \vspace{-0.2in}
    \caption{ICDAR 2024/2025 MapText Competitions Rumsey Dataset. ``Valid'' means the word is legible and fully visible.} 
    \begin{tabular}{lSSS} \toprule
                             & \textbf{Train}  & \textbf{Validation}   & \textbf{Test}  \\ \midrule
    \textbf{Tiles (images)}          & {200}             & {40}                    & {700}            \\ \cmidrule(lr){1-4} 
    \textbf{Words}                   & 34,518          & 5,544                 & 128,457        \\ 
    \textbf{Valid Words}             & 30,563          & 4,860                 & 111,821        \\ \cmidrule(lr){1-4} 
    \textbf{Label Groups}            & 21,205          & 3,502                 & 78,582         \\ 
    \textbf{Groups with $>$1 Words} & 9,772            & 1,453                 & 36,655         \\ \bottomrule    
\end{tabular}
\label{tab: data stats}
\end{table}

\section{\modelname: Generating Link Paths}\label{app: alg}
During inference, \modelname~builds text linking paths by first greedily assigning each word its most probable successor.
This can result in conflicts where multiple words point to the same successor. To resolve such ambiguities, \modelname~retains only the predecessor with the highest probability and allows others to reassign their next-best successors. Algorithm~\ref{alg: assign_succ} outlines this disambiguation process, ensuring each word has at most one valid successor or marks the end of a sequence. The resulting successor list is then used by Algorithm~\ref{alg: gen_path} to generate complete linking paths.

\begin{algorithm}
\caption{Assign Successors with Probability}
\begin{algorithmic}[1]
\Procedure{AssignSuccessors}{$words, probabilities$}
    \State Initialize $word2succ \gets \{\}$, $succ2word \gets \{\}$
    \While{$|word2succ| < |words|$}
        \For{$i \gets 0$ to $|words| - 1$}
            \State $prob \gets probabilities[i]$
            \State $j \gets \text{argmax}(prob)$
            \If{$j \notin succ2word$}
                \State $word2succ[i] \gets j$
                \State $succ2word[j] \gets i$
            \Else
                \State $old\_i \gets succ2word[j]$
                \If{$old\_i == i$} \textbf{continue}
                \ElsIf{$old\_i == j$}
                    \State $word2succ[i] \gets j$
                    \State $succ2word[j] \gets i$
                \ElsIf{$i == j$}
                    \State $word2succ[i] \gets j$
                \ElsIf{$probabilities[i][j] > probabilities[old\_i][j]$}
                    \State $word2succ[i] \gets j$
                    \State $succ2word[j] \gets i$
                    \State Remove $old\_i$ from $word2succ$
                    \State $probabilities[old\_i][j] \gets 0$
                \Else
                    \State $probabilities[i][j] \gets 0$
                \EndIf
            \EndIf
        \EndFor
    \EndWhile
    \State $successors \gets [word2succ[i] \text{ for } i \in 0 \dots |words|-1]$
    \State \Return $successors$
\EndProcedure
\end{algorithmic}
\label{alg: assign_succ}
\end{algorithm}

\begin{algorithm}
\caption{Generate Link Paths}
\begin{algorithmic}[1]
\Procedure{GeneratePaths}{$words, successors$}
    \State Initialize $word2succ \gets \{\}$, $succ2word \gets \{\}$
    \ForAll{$(word, successor)$ in $(words, successors)$}
        \State $word2succ[element] \gets successor$
        \If{$word \ne successor$}
            \State $succ2word[successor] \gets word$
        \EndIf
    \EndFor

    \State Initialize $groups \gets [\,]$, $seen \gets \{\}$

    \Function{FindPath}{$x$}
        \State $path \gets [x]$
        \State $local\_seen \gets \{\}$
        \While{$x \notin localSeen$ and $x \notin seen$}
            \If{$x$ in $succ2word$}
                \State $x \gets succ2word[x]$
                \State Insert $x$ at beginning of $path$
                \State Add $x$ to $local\_seen$
            \Else
                \State \textbf{break}
            \EndIf
        \EndWhile
        \State \Return $path$
    \EndFunction

    \ForAll{$word$ in $words$}
        \If{$word \notin seen$ and $word2succ[word] == word$}
            \State $path \gets$ \Call{FindPath}{$word$}
            \State Add $path$ to $groups$
            \State Add $each\_element$ in $path$ to $seen$
        \EndIf
    \EndFor
    \State \Return $groups$
\EndProcedure
\end{algorithmic}
\label{alg: gen_path}
\end{algorithm}


\end{document}